\documentclass{article}
\usepackage{amssymb}
\usepackage{amsmath}
\usepackage[margin=0.9in]{geometry}
\usepackage{graphicx}
\usepackage{subcaption}
\usepackage{caption}
\usepackage{booktabs}
\usepackage{algorithm}
\usepackage{algpseudocode}
\usepackage{listings}
\usepackage{xcolor}
\usepackage{url}
\usepackage{hyperref}

\hypersetup{
  colorlinks=true,
  urlcolor=blue,
  citecolor=blue,
  linkcolor=blue
}

\title{PROMA: Projected Microbatch Accumulation for Reference-Free Proximal Policy Updates}

\author{Nilin Abrahamsen}

\date{}

\begin{document}

\maketitle

\begin{abstract}
This note introduces Projected Microbatch Accumulation (PROMA), a reference-free proximal policy method that controls KL divergence by projecting away high-variance components of the policy gradient. Two variants are presented. In the accumulation-based variant, the running gradient is projected orthogonal to the sequence-wise log-probability gradients of each microbatch. In the intra-microbatch variant, a factored projection using dominant subspaces of activations and gradient outputs is applied independently within each microbatch, making it compatible with standard data-parallel training. Empirically, the accumulation variant achieves tighter per-step KL control than GRPO with PPO clipping, while the intra-microbatch variant achieves the best validation performance.
\end{abstract}

\section{Background}

Proximal Policy Optimization (PPO)~\cite{ppo} and its variants stabilize reinforcement learning by clipping the policy likelihood ratio to limit deviations from a reference policy $\pi_{\text{old}}$. GRPO~\cite{grpo,gspo} combines PPO-style clipping with a group-relative advantage estimator and has become the dominant approach for LLM fine-tuning. CISPO~\cite{CISPO} is another variant that clips the importance-sampling ratio while ensuring that all tokens contribute to the gradient. More broadly, proximal policy methods can be viewed as tractable approximations to the natural policy gradient~\cite{nat-pg}, whose exact computation is infeasible at scale.

\section{Accumulation-based PROMA}

Accumulation-based PROMA modifies the gradient during the backward pass. For each microbatch, the sequence-wise log-probability gradients ($\texttt{mcb\_seq\_grads}$ in Algorithm~\ref{alg:proma}) are computed as part of evaluating the policy gradient, denoted $\texttt{mcb\_policy\_grad}$. Before gradients are accumulated, the running gradient $\texttt{t.grad}$ of a parameter tensor $\texttt{t}$ is updated by projecting out components that lie in the span of the sequence-wise log-probability gradients of the current microbatch.

\begin{algorithm}[H]
    \centering
\begin{lstlisting}[language=Python]
def PROMA_accumulate(t, mcb_policy_grad, mcb_seq_grads)
    """
    k: number of sequences in microbatch
    t: (d,) tensor of model weights in the layer
    mcb_policy_grad: (d,) vanilla policy gradient w.r.t. tensor t for the microbatch
    mcb_seq_grads: (d, k), sequence-wise gradients of log_prob.sum() w.r.t tensor t
    """
    complemented_acc_grad = project_to_complement(t.grad, mcb_seq_grads)
    t.grad = complemented_acc_grad + mcb_policy_grad

def project_to_complement(acc_vec, vecs):
    Q, _ = torch.linalg.qr(vecs, mode="reduced")  # Q: (d, k)
    return acc_vec - Q @ (Q.T @ acc_vec)

\end{lstlisting}
    \caption{$\texttt{PROMA\_accumulate}$ is called during the backward pass of the policy gradient loss and modifies the gradient accumulation.
    Before adding the microbatch gradient,
    the partially accumulated gradient is projected to be orthogonal to the gradient of each sequence in the current microbatch ($\texttt{mcb\_seq\_grads}$). }
    \label{alg:proma}
\end{algorithm}
The parameter update $d\theta$ incurs a KL divergence that grows with the squared overlaps $\mathbb E_s[(d\theta\;\cdot\;\nabla_\theta\log\pi(s))^2]$, where $\nabla\log\pi(s)$ is the direction of fastest probability change for sequence $s$. PROMA controls this by projecting away these components of the gradient update.

\paragraph{Efficiency}
To exactly project away the $k$ sequences in the microbatch takes $\approx 2k^2d$ FLOPs where $d$ is the number of model parameters. If the microbatch size is large then this can be applied group-wise so that $k\sim8$. The cost $2k^2d$ of PROMA is then negligible compared to the $T\times d$ operations required to form the gradient, where $T$ is the number of tokens. Another option is to apply an approximate iterative projection which takes $O(kd)$ operations.

\section{Intra-microbatch PROMA}
\label{sec:intra}

While PROMA modifies gradient accumulation across microbatches, the Intra-PROMA method defined below instead acts within each microbatch. Because the projection depends only on the current microbatch, Intra-PROMA is embarrassingly parallel across microbatches and compatible with standard data-parallel training.

\begin{algorithm}[H]
    \centering
\begin{lstlisting}[language=Python]
def proma_intra(token_advantages, act_in, grad_out, r=100, shrinkage=1.0):
    """
    Project out gradient components in the subspace spanned by
    top-r singular vectors of activations and gradient outputs.
    """
    r = min(r, act_in.shape[0], act_in.shape[1], grad_out.shape[1])

    Q_a = approx_rank_r_basis(act_in, r)   # (d_in, r)
    Q_g = approx_rank_r_basis(grad_out, r) # (d_out, r)

    policy_grad = (token_advantages * grad_out.T) @ act_in / len(token_advantages)
    projection = Q_g @ (Q_g.T @ policy_grad @ Q_a) @ Q_a.T
    return policy_grad - shrinkage * projection

def approx_rank_r_basis(X, r, power_iters=1):
    """Approximate top-r right singular vectors via randomized SVD."""
    n = X.shape[0]
    omega = torch.randn(n, r)
    Y = X.T @ omega
    for _ in range(power_iters):
        Y = X.T @ (X @ Y)
    Q, _ = torch.linalg.qr(Y)
    return Q
\end{lstlisting}

    \caption{$\texttt{proma\_intra}$ computes the Intra-PROMA gradient for a single layer within a microbatch. The projection subtracts gradient components lying in the dominant subspaces of activations and gradient outputs, approximated via randomized SVD.}
    \label{alg:intra}
\end{algorithm}

The factored subspace approximation is similar to the Kronecker factorizations underlying K-FAC~\cite{kfac} and related second-order optimizers~\cite{ACKTR}, which apply a sandwich product $G^{-1}(\cdot)A^{-1}$ to precondition the gradient. In contrast to K-FAC, PROMA applies a sandwich projection to the subtracted component:
\begin{equation}\label{eq:intersect}
\texttt{grad} - Q_g Q_g^\top \texttt{grad}\, Q_a Q_a^\top,
\end{equation}
meaning that intra-PROMA removes the \emph{intersection} of the leading row and column substraces to control KL without impeding training progress. Alternatively one could apply a sandwich product of projections to the gradient~\eqref{eq:kfaclike} similarly to the $G^{-1}(\cdot)A^{-1}$ in K-FAC:
\begin{equation}\label{eq:kfaclike}
(I - Q_g Q_g^\top)\,\texttt{grad}\,(I - Q_a Q_a^\top).
\end{equation}
\eqref{eq:kfaclike} would correspond to removing the \emph{union} of leading row and column subspaces, and I found that this hurt validation performance compared to \eqref{eq:intersect}.

\section{Results}
\label{sec:results}

\paragraph{Experimental setup}
The demonstration was implemented in VeRL~\cite{verl}
(\url{https://github.com/nilin/proma}).
The model is trained on MBPP~\cite{mbpp} and validated on HumanEval~\cite{humaneval}, two related but distinct code generation benchmarks, using the Qwen-3 0.6B model~\cite{qwen3_2025}.
The training batch consists of 64 prompts, each generating $n=16$ rollout samples. Policy updates use a mini-batch size of 32 and micro-batch size of 8, yielding $k=4$ microbatches per gradient accumulation step. All methods use a single PPO epoch per batch, the AdamW optimizer with weight decay $0.01$ and gradient clipping at $1.0$, and a constant learning rate schedule. Learning rates are swept over $\{2\times10^{-6},\; 5\times10^{-6},\; 1\times10^{-5}\}$. The maximum prompt length is 512 tokens and the maximum response length is 1024 tokens. No KL penalty is applied in any configuration. Validation is performed every 2 training steps for 20 steps.
All methods use the same GRPO group-relative advantage estimator~\cite{grpo}.

\paragraph{Baselines and naming}
This experiment compares four configurations.
\textbf{GRPO (w/ PPO-clip)} uses PPO-style clipping of the likelihood ratio with clip ratio $\varepsilon=0.2$.
\textbf{GRPO (no PPO-clip)} uses the same group-relative advantage but without clipping the likelihood ratio.
\textbf{PROMA (intra)} applies Intra-PROMA (Algorithm~\ref{alg:intra}) with projection dimension $d=10$.
\textbf{PROMA (accumulation)} applies the accumulation-based PROMA (Algorithm~\ref{alg:proma}) on the last 2 out of 4 microbatches.
The PROMA configurations do not use PPO clipping, isolating the effect of the projected accumulation.

Two validation metrics are reported: the \emph{best} validation score (peak score over the course of each run) and the \emph{final} validation score (score at the last step), both aggregated as the median across runs.

\begin{figure}[H]
    \centering
    \begin{subfigure}{0.48\linewidth}
        \centering
        \includegraphics[width=\linewidth]{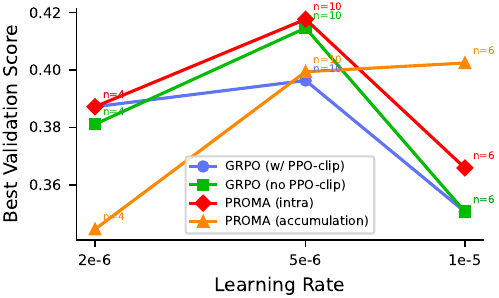}
        \caption{Best validation score vs.\ learning rate}
        \label{fig:val:lr-best}
    \end{subfigure}
    \hfill
    \begin{subfigure}{0.48\linewidth}
        \centering
        \includegraphics[width=\linewidth]{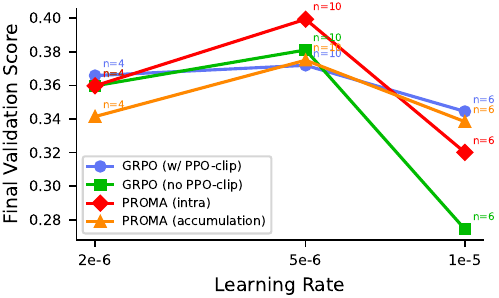}
        \caption{Final validation score vs.\ learning rate}
        \label{fig:val:lr-last}
    \end{subfigure}
    \begin{subfigure}{0.48\linewidth}
        \centering
        \includegraphics[width=\linewidth]{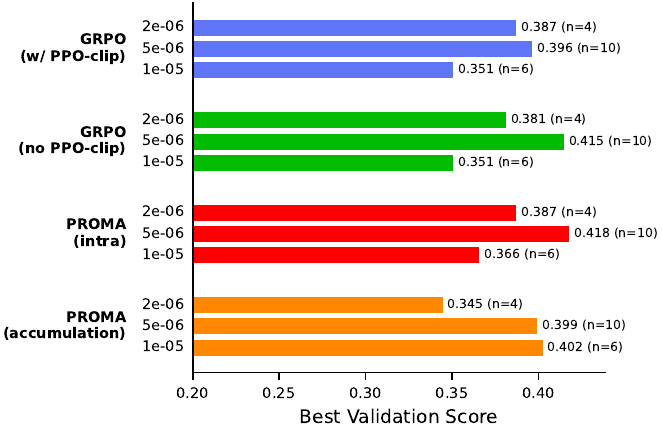}
        \caption{Best validation score per learning rate}
        \label{fig:val:bar-best}
    \end{subfigure}
    \hfill
    \begin{subfigure}{0.48\linewidth}
        \centering
        \includegraphics[width=\linewidth]{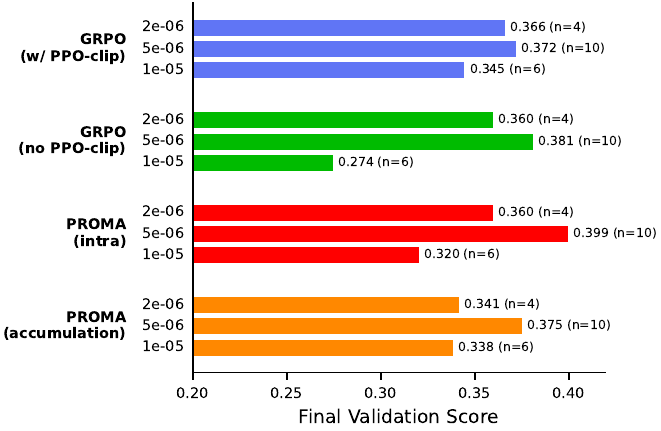}
        \caption{Final validation score per learning rate}
        \label{fig:val:bar-last}
    \end{subfigure}

    \caption{MBPP $\to$ HumanEval validation results (Qwen-3 0.6B). (a--b) Learning rate sensitivity; each marker shows the median across seeds, $n$ indicates number of runs. (c--d) Per-learning-rate breakdown; each bar shows the median, $n$ denotes the number of seeds.}
    \label{fig:val}
\end{figure}

\begin{figure}[H]
    \centering
    \begin{subfigure}{0.45\linewidth}
        \centering
        \includegraphics[width=\linewidth]{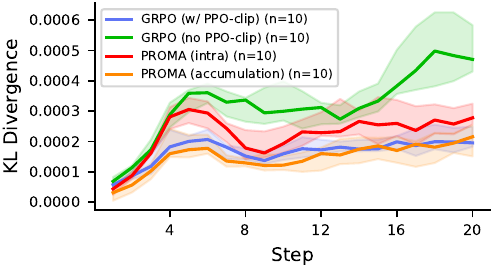}
        \caption{Per-step KL divergence (smoothed)}
        \label{fig:kl:ppo-kl}
    \end{subfigure}
    \hfill
    \begin{subfigure}{0.45\linewidth}
        \centering
        \includegraphics[width=\linewidth]{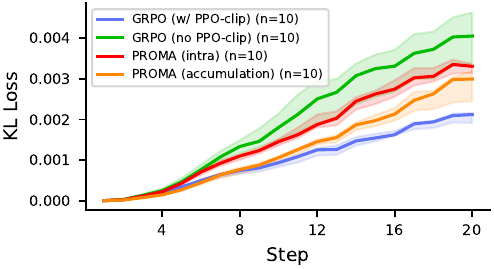}
        \caption{KL divergence from initial policy}
        \label{fig:kl:kl-loss}
    \end{subfigure}
    \caption{KL metrics at lr $= 5\times10^{-6}$. Lines show median, shaded regions show IQR.}
    \label{fig:kl}
\end{figure}

Figure~\ref{fig:val}a--b shows learning rate sensitivity. PROMA (intra) achieves the highest best validation score at lr $= 5\times10^{-6}$, and PROMA (accumulation) is the most robust to higher learning rates, maintaining strong performance at lr $= 1\times10^{-5}$ where both GRPO variants degrade. For final validation scores, GRPO (no PPO-clip) drops sharply at lr $= 1\times10^{-5}$, while GRPO (w/ PPO-clip) and PROMA (accumulation) remain more stable. Figure~\ref{fig:val}c--d provides a per-learning-rate breakdown. 

Figure~\ref{fig:kl} shows KL dynamics at lr $= 5\times10^{-6}$. GRPO (no PPO-clip) exhibits the largest per-step KL divergence (Figure~\ref{fig:kl}a), while both PROMA variants maintain tighter local KL control, comparable to GRPO (w/ PPO-clip). PROMA (accumulation) achieves the lowest per-step KL overall. In terms of cumulative drift from the initial policy (Figure~\ref{fig:kl}b), GRPO (w/ PPO-clip) stays closest to the initial policy, with PROMA variants in between, drifting less than GRPO (no PPO-clip) while still allowing more policy movement than the clipped baseline. PROMA (intra) achieves the best validation performance across all methods, while PROMA (accumulation) provides the tightest local KL control, both without relying on likelihood ratio clipping.

\section{Discussion}

PROMA is a followup to ISOPO~\cite{isopo} which likewise aimed to define a proximal gradient update without a reference policy. In contrast to ISOPO which explored pre-conditioning based on re-scaling, PROMA instead projects away noisy directions to achieve proximal policy updates. The usefulness of projecting the accumulated gradient is consistent with findings in computational physics in which the SPRING optimizer~\cite{spring} applied a projected momentum scheme to the MinSR~\cite{minsr} optimizer to achieve increased accuracy. Both MinSR and SPRING are variants of the variational Monte Carlo (VMC) method~\cite{vmc} for molecular energy calculation, a method which is formally similar to policy optimization. This parallel highlights a shared geometric structure underlying stochastic optimization methods in both computational physics and reinforcement learning.

\section{Acknowledgements}

I thank Molei Tao and Michael Luo for inspiring conversations and encouragement.

\bibliographystyle{abbrv}
\bibliography{bibliography}
\end{document}